\documentclass[10pt,twocolumn,letterpaper]{article}

\usepackage{iccv}
\usepackage{times}
\usepackage{epsfig}
\usepackage{graphicx}
\usepackage{amsmath}
\usepackage{amssymb}

\usepackage[ruled,vlined]{algorithm2e}
\usepackage{makecell}

\DeclareMathOperator*{\argmax}{argmax}

\usepackage{xcolor,colortbl}
\definecolor{Gray}{gray}{0.9}

\usepackage[breaklinks=true,bookmarks=false]{hyperref}

\iccvfinalcopy 

\def\iccvPaperID{383} 

\ificcvfinal\fi

\begin{document}

\title{SegSort: Segmentation by Discriminative Sorting of Segments}

\author{
Jyh-Jing Hwang$^{1,2}$ \quad Stella X. Yu$^1$ \quad Jianbo Shi$^2$\\ 
Maxwell D. Collins$^3$ \quad Tien-Ju Yang$^4$ \quad Xiao Zhang$^3$ \quad Liang-Chieh Chen$^3$ \\
$^1$UC Berkeley / ICSI \quad $^2$University of Pennsylvania \quad $^3$Google Research \quad $^4$MIT\\
{\tt \small \{jyh,stellayu\}@berkeley.edu  \quad \{jyh,jshi\}@seas.upenn.edu} \\
{\tt \small \{maxwellcollins,tjy,andypassion,lcchen\}@google.com}
}



\maketitle
\ificcvfinal\thispagestyle{empty}\fi

\maketitle


\begin{abstract}
Almost all existing deep learning approaches for semantic segmentation tackle this task as a pixel-wise classification problem.
Yet humans understand a scene not in terms of pixels, but by decomposing it into perceptual groups and structures that are the basic building blocks of recognition.
This motivates us to propose an end-to-end pixel-wise metric learning approach that mimics this process.
In our approach, the optimal visual representation determines the right segmentation within individual images and associates segments with the same semantic classes across images.
The core visual learning problem is therefore to maximize the similarity within segments and minimize  the similarity between segments.
Given a model trained this way, inference is performed consistently by extracting pixel-wise embeddings and clustering, with the semantic label determined by the majority vote of its nearest neighbors from an annotated set.
As a result, we present the SegSort, as a first attempt using deep learning for unsupervised semantic segmentation, achieving $76\%$ performance of its supervised counterpart.
When supervision is available, SegSort shows consistent improvements over conventional approaches based on pixel-wise softmax training.
Additionally, our approach produces more precise boundaries and consistent region predictions.
The proposed SegSort further produces an interpretable result, as each choice of label can be easily understood from the retrieved nearest segments.

\end{abstract}

\section{Introduction}
\label{sec:intro}

\begin{figure}[t]
    \centering
    \includegraphics[width=1.0\linewidth]{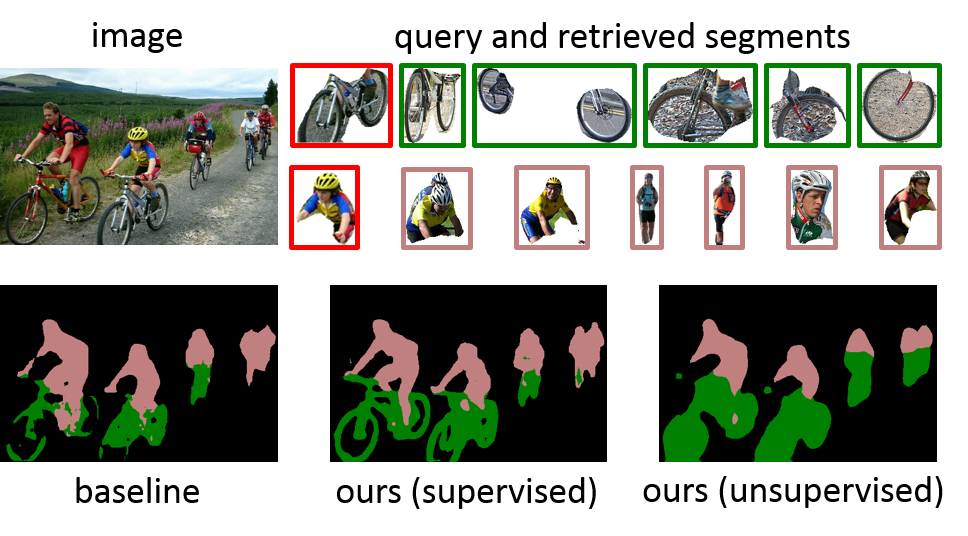}
    \caption{
Top: Our proposed approach partitions an image in the embedding space into aligned segments (framed in red) and assign the majority labels from retrieved segments (framed in green or pink). Bottom: Our approach presents the first deep learning based unsupervised semantic segmentation (right). If supervised, our approach produces more consistent region predictions and precise boundaries in the supervised setting (middle) compared to its parametric counterpart (left).
}
    \label{fig:teaser}
\end{figure}

%
%
%

%
%
%


%


Semantic segmentation is usually approached by extending image-wise classification \cite{lecun1989backpropagation,krizhevsky2012imagenet} to pixel-wise classification, deployed in a fully convolutional fashion ~\cite{long2015fully}.
In contrast, we study the semantic segmentation task in terms of perceiving an image in groups of pixels and associating objects from a large set of images.
Particularly, we take the perceptual organization view \cite{tenenbaum1983role, biederman1987recognition} that pixels group by visual similarity and objects form by visual familiarity; consequently a representation is developed to best relate pixels and segments to each other in the visual world.
Our method, such motivated, not only achieves better supervised semantic segmentation but also presents the first attempt using deep learning for {\it unsupervised} semantic segmentation.


We formulate this intuition as an end-to-end metric learning problem.
Each pixel in an image is mapped via a CNN to a point in some visual embedding space, and nearby points in that space indicate pixels belonging to the same segments.
From all the segments collected across images, clusters in the embedding space form semantic concepts.
In other words, we {\it sort segments} with respect to their visual and semantic attributes.
The optimal visual representation delivers the right segmentation within individual images and associates segments with the same semantic classes across images, yielding a non-parametric model as its complexity scales with number of segments (exemplars).

We derive our method based on maximum likelihood estimation of a single equation, resulting in a two-stage Expectation-Maximization (EM) framework. The first stage performs a spherical (von Mises-Fisher) K-Means clustering~\cite{banerjee2005clustering} for image segmentation. The second stage adapts the E-step for a pixel-to-segment loss to optimize the metric learning CNN.


As a result, we present the SegSort (Segment Sorting) as a first attempt to apply deep learning for semantic segmentation from the {\it unsupervised} perspective. Specifically, we create pseudo segmentation masks aligned with visual cues using a contour detector \cite{arbelaez2011contour,hwang2015pixel,xie2015holistically} and train the pixel-wise embedding network to separate all the segments. The unsupervised SegSort achieves $76\%$ performance of its supervised counterpart. We further show that various visual groups are automatically discovered in our framework.

When supervision is available (\ie, supervised semantic segmentation), we segment each image with the spherical K-Means clustering and train the network following the same optimization, but incorporated with Neighborhood Components Analysis criterion \cite{goldberger2005neighbourhood,wu2018improving} for semantic labels.

To summarize our major contributions:
\vspace{-6pt}
\begin{enumerate}
\setlength\itemsep{-2pt}

\item We present the first end-to-end trained non-parametric approach for supervised semantic segmentation, with performance exceeding its parametric counterparts that are trained with pixel-wise softmax loss.

\item We propose the first unsupervised deep learning approach for semantic segmentation, which achieves $76\%$ performance of its supervised counterpart.

\item Our segmentation results can be easily understood from retrieved nearest segments and readily interpretable.

\item Our approach produces more precise boundaries and more consistent region segmentations compared with parametric pixel-wise prediction approaches.

\item We demonstrate the effectiveness of our method on two challenging datasets, PASCAL VOC 2012~\cite{pascal-voc-2012} and Cityscapes~\cite{cordts2016cityscapes}.

\end{enumerate}

\section{Related Works}
\label{sec:work}

\noindent \textbf{Segmentation and Clustering.}
Segmentation involves extracting representations from local patches and clustering them based on different criteria, \eg, fitting mixture models \cite{yang2008unsupervised,belongie1998color}, mode-finding \cite{comaniciu2002mean,banerjee2005clustering}, or graph partitioning \cite{felzenszwalb2004efficient,shi2000normalized,malik2001contour,stella2003multiclass,yu2004segmentation}. The mode-finding algorithms, \eg, mean shift \cite{comaniciu2002mean} or K-Means \cite{hartigan1979algorithm,banerjee2005clustering}, are mostly related. Traditionally, pixels are encoded in a joint spatial-range domain by a single vector with their spatial coordinates and visual features concatenated. Applying mean shift or K-Means filtering can thus converge for each pixel. Spectral graph theory \cite{chung1997spectral}, and in particular the Normalized Cut \cite{shi2000normalized} criterion provides a way to further integrate global image information for better segmentation. More recently, superpixel approaches \cite{achanta2012slic} emerge to be a popular pre-processing step that helps reduce the computation, or can be used to refine the semantic segmentation predictions \cite{gadde2016superpixel}. However, the challenge of perceptual organization is to process information from different levels together to form consensus segmentation. Hence, our proposed approach aims to integrate image segmentation and clustering into end-to-end embedding learning for semantic segmentation.

\noindent \textbf{Semantic Segmentation.}
Current state-of-the-art semantic segmentation models are based on Fully Convolutional Networks \cite{lecun1989backpropagation,sermanet2013overfeat,long2015fully}, tackling the problem via pixel-wise classification. Given limited local context, it may be ambiguous to correctly classify a single pixel, and thus it is common to resort to multi-scale context information \cite{ he2004multiscale,shotton2009textonboost,kohli2009robust,ladicky2009associative,gould2009decomposing, yao2012describing, mostajabi2014feedforward,aaf2018,hwang2019adversarial}. Typical approaches include image pyramids \cite{farabet2013learning, pinheiro2014recurrent, eigen2015predicting, lin2015efficient, chen2015attention, chen2016deeplab} and encoder-decoder structures \cite{badrinarayanan2015segnet, ronneberger2015u, lin2016refinenet, fu2017stacked, peng2017large, yu2018learning, zhang2018exfuse, deeplabv3plus2018}. Notably, to better capture multi-scale context, PSPNet \cite{zhao2016pyramid} performs spatial pyramid pooling \cite{grauman2005pyramid,lazebnik2006beyond,liu2015parsenet} at several grid scales, while DeepLab \cite{chen2016deeplab,chen2017rethinking,yang2019deeperlab} applies the ASPP module (Atrous Spatial Pyramid Pooling) consisting of several parallel atrous convolution \cite{holschneider1989real,giusti2013fast,sermanet2013overfeat,papandreou2014untangling} with different rates. In this work, we experiment with applying our proposed training algorithm to PSPNet and DeepLabv3+, and show consistent improvements.

Before deep learning takes a leap, non-parametric methods for semantic segmentation are explored. In the unsupervised setting,  \cite{russell2009segmenting} proposes data-driven boundary and image grouping,  formulated with MRF to enhance semantic boundaries; \cite{tighe2010superparsing} extracts superpixels before nearest neighbor search;  \cite{liu2011nonparametric} performs dense SIFT to find dense deformation fields between images to segment and recognize a query image. With supervision, \cite{malisiewicz2008recognition} learns semantic object exemplars for detection and segmentation.

It is worth noting Kong and Fowlkes \cite{kong2018recurrent} also integrate vMF mean-shift clustering into the semantic segmentation pipeline. However, the clustering with contrastive loss is used for regularizing features and the whole system still relies on softmax loss to produce the final segmentation.

Our work also bears a similarity to the work Scene Collaging \cite{isola2013scene}, which presents a nonparametric scene grammar for parsing the images into segments for which object labels are retrieved from a large dataset of example images. 

\begin{figure*}[t]
    \centering
    \includegraphics[width=1.0\textwidth]{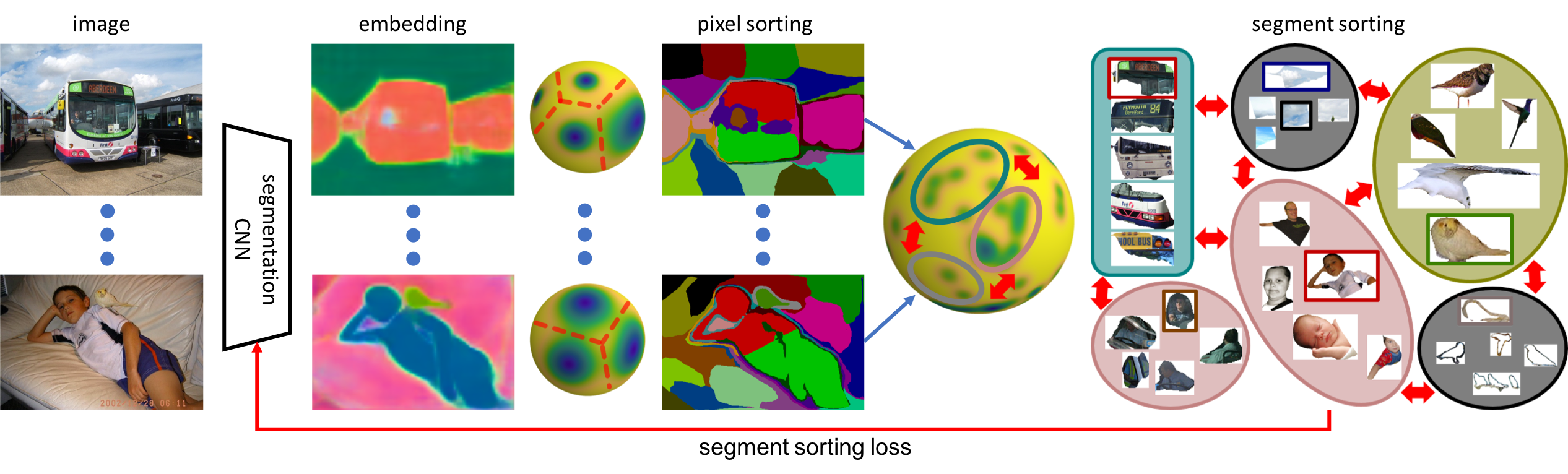}
    \caption{The overall training diagram for our proposed framework, Segment Sorting (SegSort), with the vMF clustering \cite{banerjee2005clustering}. Given a batch of images (leftmost), we compute pixel-wise embeddings (middle left) from a metric learning segmentation network. Then we segment each image with the vMF clustering (middle right), dubbed pixel sorting. We train the network via the maximum likelihood estimation derived from a mixture of vMF distributions, dubbed segment sorting. In between, we also illustrate how to process pixel-wise features on a hyper-sphere for pixel and segment sorting. A segment (rightmost) is color-framed with its corresponding vMF clustering color if in the displayed images. Unframed segments from different images are associated in the embedding space. The inference is done with the same procedure but using the k-nearest neighbor search to associate segments in the training set.}
    \label{fig:main}
\end{figure*}

\noindent \textbf{Metric Learning.}
Metric learning approaches \cite{koestinger2012large,goldberger2005neighbourhood} have achieved remarkable performance on different vision tasks, such as image retrieval \cite{wu2017sampling,wu2018unsupervised,wu2018improving} and face recognition \cite{taigman2014deepface,wang2018cosface,schroff2015facenet}. Such tasks usually involve open world recognition, since classes during testing might be disjoint from the ones in the training set. Metric learning minimizes intra-class variations and maximizes inter-class variations with pairwise losses \eg, contrastive loss \cite{bromley1994signature} and triplet loss \cite{hoffer2015deep}. Recently, Wu \etal \cite{wu2018unsupervised} propose a non-parametric softmax formulation for training feature embeddings to separate every image for unsupervised image recognition and retrieval. The non-parametric softmax is further incorporated with Neighborhood Components Analysis \cite{goldberger2005neighbourhood} to improve generalization for supervised image recognition \cite{wu2018improving}. An important technical point on metric learning is normalization \cite{wang2017normface,schroff2015facenet} so that features lie on a hypersphere, which is why the vMF distribution is of particular interest.

\section{Method}

Our end-to-end learning framework consists of three sequential components:
1) A CNN, \eg,  DeepLab~\cite{deeplabv3plus2018}, FCN~\cite{long2015fully}, or PSPNet~\cite{zhao2016pyramid}, that generates pixel-wise embeddings from an image.
2) A clustering method that partitions the pixel-wise embeddings into a fine segmentation, dubbed pixel sorting.
3) A metric learning formulation for separating and grouping the segments into semantic clusters, dubbed segment sorting.

We start with an assumption that the pixel-wise normalized embeddings from the CNN within a segment follow a von Mises-Fisher (vMF) distribution.
We thus formulate the pixel sorting with spherical K-Means clustering and the segment sorting with corresponding maximum likelihood estimation.
During inference, the segment sorting is replaced with k-nearest neighbor search.
We then apply to each query segment the majority label of retrieved segments.

We now give a high level mathematical explanation of the entire optimization process.
Let $\mathcal{V}=\{\pmb{v}_i\}=\{\phi(x_i)\}$ be the set of pixel embeddings where $\pmb{v}_i$ is produced by a CNN $\phi$ centered at pixel $x_i$. Let $\mathcal{Z}=\{z_i\}$ be the image segmentation with $k$ segments, or $z_i=s$ indicates if a pixel $i$ belongs to a segment $s$. Let $\Theta=\{\theta_{z_i}\}$ be the set of parameters that capture the representative feature of a segment through a predefined distribution $f$ (mixture of vMF here).
Our main optimization objective can be concluded as:
\begin{equation}
\label{eqn:one}
\min_{\phi, \mathcal{Z}, \Theta} -\log P(\mathcal{V}, \mathcal{Z}\mid \Theta) = \min_{\phi, \mathcal{Z}, \Theta} -\sum_i \log \frac{1}{k} f_{z_i}(\pmb{v}_i\mid \theta_{z_i}).
\end{equation}
In pixel sorting, we use a standard EM framework to find the optimal $\mathcal{Z}$ and $\Theta$, with $\phi$ fixed. In segment sorting, we adapt the previous E step for loss calculation through a set of images to optimize $\phi$, with $\mathcal{Z}$ and $\Theta$ fixed. Performing pixel sorting and segment sorting can thus be viewed as a two-stage EM framework.

This section is organized as follows.
We first describe the pixel sorting in Sec.~\ref{sec:clustering}, which includes a brief review of spherical K-Means clustering and creation of aligned segments.
We then derive two forms of the segment sorting loss for segment sorting in Sec.~\ref{sec:training}.
Finally, we describe the inference procedure in Sec.~\ref{sec:inference}.
The overall training diagram is illustrated in Fig.~\ref{fig:main} and the summarized algorithm can be found in the supplementary.

\subsection{Pixel Sorting}
\label{sec:clustering}

We briefly review the vMF distribution and its corresponding spherical K-Means clustering algorithm~\cite{banerjee2005clustering}, which is used to segment an image as pixel sorting.

We assume the pixel-wise $d$-dimensional embeddings $\pmb{v}\in \mathbb{S}^{d-1}$ (CNN's last layer features after normalization) within a segment follow a vMF distribution.
vMF distributions are of particular interest as it is one of the simplest distributions with properties analogous to those of the multivariate Gaussian for directional data.
Its probability density function is given by
\begin{equation}
\label{eqn:vmf}
f(\pmb{v} \mid \pmb{\mu}, \kappa) = C_d(\kappa)
\exp(\kappa \pmb{\mu}^\top \pmb{v}),
\end{equation}
where $C_d(\kappa) = \frac{ \kappa^{d/2-1} }{ (2\pi)^{d/2} I_{d/2-1}(\kappa) }$
is the normalizing constant where $I_r(\cdot)$ represents the modified Bessel function of the first kind and order $r$.
$\pmb{\mu}=\sum_i \pmb{v}_i / ||\sum_i \pmb{v}_i||$ is the mean direction and $\kappa \ge 0$ is the concentration parameter.
Larger $\kappa$ indicates stronger concentration about $\mu$.
In our particular case, we assume a constant $\kappa$ for all vMF distributions to circumvent the expensive calculation of $C_d(\kappa)$.

The embeddings of an image with $k$ segments can thus be considered as a mixture of $k$ vMF distributions with a uniform prior, or
\begin{equation}
\label{eqn:movmf}
f(\pmb{v} \mid \Theta) = \sum_{s=1}^k \frac{1}{k} f_s(\pmb{v} \mid \pmb{\mu}_s, \kappa),
\end{equation}
where $\Theta = \{\pmb{\mu}_1, \cdots, \pmb{\mu}_k, \kappa\}$. Let $z_i$ be the hidden variable that indicates a pixel embedding $\pmb{v}_i$ belongs to a particular segment $s$, or $z_i=s$. Let $\mathcal{V}=\{\pmb{v}_1, \cdots, \pmb{v}_k\}$ be the set of pixel embeddings and $\mathcal{Z}=\{z_1, \cdots, z_k\}$ be the set of corresponding hidden variables. The log-likelihood of the observed data is thus given by
\begin{equation}
\label{eqn:data_likelihood}
\log P(\mathcal{V}, \mathcal{Z}\mid \Theta) = \sum_i \log \frac{1}{k} f_{z_i}(\pmb{v}_i\mid \pmb{\mu}_{z_i},\kappa).
\end{equation}
Since $\mathcal{Z}$ is unknown, the EM framework is used to estimate this otherwise intractable maximum likelihood, resulting in the spherical K-Means algorithm with an assumption of $\kappa \mapsto \infty$.  This assumption holds if all the embeddings within a region are the same (homogeneous), which will be our training objective described in Sec.~\ref{sec:training}.

The E-step that maximizes the likelihood of Eqn.~\ref{eqn:data_likelihood} is to assign $z_i=s$ with a posterior probability~\cite{neal1998view}:
\begin{equation}
\label{eqn:posterior}
    p(z_i=s|\pmb{v}_i, \Theta) = \frac{f_s(\pmb{v}_i \mid \Theta)}{\sum_{l=1}^k f_l(\pmb{v}_l|\Theta)}.
\end{equation}
In the setting of K-Means, we use hard assignments to update $z_i$, or $z_i = \argmax_{s} p(z_i=s\mid \pmb{v}_i, \Theta)=\argmax_{s} \pmb{\mu}_{s}^\top \pmb{v}_i$. We further denote the set of pixels within a segment $c$ as $\mathcal{R}_c$; hence $p(z_i=c\mid \pmb{v}_i, \Theta)=1$ if $i\in \mathcal{R}_c$ or $0$ otherwise after hard assignments.

The M-step that maximizes the expectation of Eqn.~\ref{eqn:data_likelihood} can be derived~\cite{banerjee2005clustering} as
\begin{equation}
\label{eqn:vmf_mu}
\hat{\pmb{\mu}}_c = \frac{\sum_i \pmb{v}_i p(z_i=c\mid \pmb{v}_i, \Theta)}{||\sum_i \pmb{v}_i p(z_i=c\mid \pmb{v}_i, \Theta)||} = \frac{\sum_{i \in \mathcal{R}_c} \pmb{v}_i}{|| \sum_{i \in \mathcal{R}_c} \pmb{v}_i ||},
\end{equation}
which is the mean direction of pixel embeddings within segment $c$.
The spherical K-Means clustering is thus done through alternating updates of $\mathcal{Z}$ (E-step) and $\Theta$ (M-step).


One problem of K-Means clustering is the dynamic number of EM steps, which would cause uncertain memory consumption during training.
However, we find in practice a small fixed number of EM steps, \ie, $10$ iterations, can already produce good segmentations.

If we only use embedding features for K-Means clustering, each resulted cluster is often disconnected and scattered.
As our goal is to spatially segment an image, we concatenate pixel coordinates with the embeddings so that the K-Means clustering is guided by spatiality. \\

\noindent {\bf Creating Aligned Segments.} Segments that are aligned with different visual cues are critical for producing coherent boundaries. However, segments produced by K-Means clustering do not always conform to the ground truth boundaries.
If one segment contains different semantic labels, it clearly contradicts our assumption of homogeneous embeddings within a segment.
Therefore, we partition a segment given the ground truth mask in the supervised setting so that each segment contains exactly a single semantic label as illustrated in Fig. \ref{fig:align}.

It is easy to see that the segments after partition are always aligned with semantic boundaries.
Furthermore, this partition creates small segments of false positives and false negatives which can naturally serve as hard negative examples during loss calculation.

\subsection{Segment Sorting}
\label{sec:training}

\begin{figure}[t]
    \centering
    \includegraphics[width=1.0\linewidth]{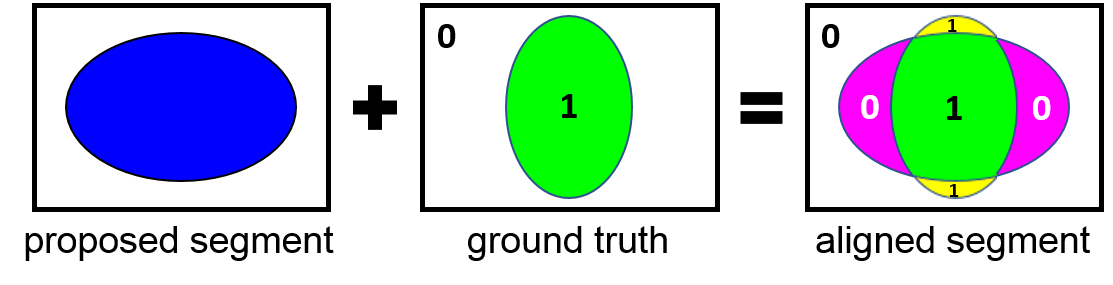}
    \caption{During supervised training, we partition the proposed segments (left) given the ground truth mask (middle). The yielded segments (right) are thus aligned with ground truth mask. Each aligned segment is labeled ($0$ or $1$) according to ground truth mask. Note that the purple and yellow segments become, respectively, false positive and false negative that help regularize predicted boundaries.}
    \label{fig:align}
\end{figure}

Following our assumption of homogeneous embeddings per segment, the training is therefore to enforce this criterion, which is done by optimizing the CNN parameters for better feature extraction.

We first define a {\it prototype} as the most representative embedding feature of a segment. Since the embeddings in a segment follow a vMF distribution, the mean direction vector $\pmb{\mu}_c$ in Eqn.~\ref{eqn:vmf_mu} can naturally be used as the prototype.

In Sec.~\ref{sec:clustering}, we consider 
the posterior probability of a pixel embedding $\pmb{v}_i$ belonging to a segment $s$ with fixed CNN parameters in Eqn.~\ref{eqn:posterior}. Now we revisit it with free CNN parameters $\phi$ and a constant hyperparameter $\kappa$:
\begin{equation}
\label{eqn:posterior_cnn}
p_\phi(z_i=s \mid \pmb{v}_i, \Theta) =  \frac{f_s(\pmb{v}_i|\Theta)}{\sum_{l=1}^k f_l(\pmb{v}_l|\Theta)} = \frac{ \exp(\kappa \pmb{\mu}_s^\top \pmb{v}_i) }{ \sum_{l=1}^k \exp(\kappa \pmb{\mu}_l^\top \pmb{v}_i) }.
\end{equation}
As both the embedding $\pmb{v}$ and prototype $\pmb{\mu}$ are of unit length, the dot product $\pmb{v}^\top \pmb{\mu} = \frac{\pmb{v}^\top \pmb{\mu}}{||\pmb{v}||\ ||\pmb{\mu}||}$ becomes the cosine similarity.
The numerator indicates the exponential cosine similarity between a pixel embedding $\pmb{v}_i$ and a particular segment prototype $\pmb{\mu}_s$.
The denominator includes the exponential cosine similarities w.r.t. all the segment prototypes.
The value of $p_\phi$ indicates the ratio of pixel embedding $\pmb{v}_i$ close to segment $s$ compared to all the other segments.

The training objective is thus to maximize the posterior probability of a pixel embedding belonging to its corresponding segment $c$ obtained from the K-Means clustering.
In other words, we want to minimize the following negative log-likelihood, or the vMF loss:
\begin{equation}
\label{eqn:vmf_loss}
    L_\text{vMF}^i = -\log p_\phi(c\mid \pmb{v}_i, \Theta) = -\log \frac{ \exp(\kappa \pmb{\mu}_c^\top \pmb{v}_{i} ) }{ \sum_{l=1}^k \exp(\kappa \pmb{\mu}_l^\top \pmb{v}_i) }.
\end{equation}
The total loss is the average over all pixels.
As a result, minimizing $L_\text{vMF}$ has two effects:
One is expressed by the numerator, where it encourages each pixel embedding to be close to its own segment prototype.
The other is from the denominator, where it encourages each embedding feature to be far away from all other segment prototypes.

Note that this vMF loss does not require any ground truth semantic labels.
We can therefore use this loss to train the CNN in an unsupervised setting.
As the loss pushes every segment as far away as possible, visually similar segments are forced to stay closer on the hypersphere.

To make use of ground truth semantic information, we consider soft neighborhood assignments in the Neighborhood Components Analysis \cite{goldberger2005neighbourhood}.
The idea of soft neighborhood assignments is to encourage the probability of one example selecting its neighbors (excluding itself) of the same category.
In our case, we want to encourage the probability of a pixel embedding $\pmb{v}_i$ selecting any other segment in the same category, denoted as  $c^+$, as its neighbors.
We can define such probability as follows, adapted from Eqn.~\ref{eqn:posterior_cnn}:
\begin{align}
\nonumber &p_\phi'(z_i=c^+\mid \pmb{v}_i, \Theta) = \frac{f_{c^+}(\pmb{v}_i|\Theta)}{\sum_{l \neq c} f_l(\pmb{v}_l|\Theta)} = \frac{ \exp(\kappa \pmb{\mu}_{c^+}^\top \pmb{v}_i) }{ \sum_{l\neq c} \exp(\kappa \pmb{\mu}_l^\top \pmb{v}_i) }, \\
&p_\phi'(z_i=c\mid \pmb{v}_i, \Theta) = 0.
\end{align}
We denote the set of segments $\{c^+\}$ w.r.t. pixel $i$ as $C_i^+$.

Our final loss function is therefore the negative log total probability of pixel $i$ selecting a neighbor prototype in the same category:
\begin{align}
\label{eqn:vmfn_loss}
\nonumber L_{\text{vMF-N}}^i =& -\log \sum_{s\in C_i^+} p_\phi'(z_i=s\mid \pmb{v}_i, \Theta) \\
=& -\log \frac{ \sum_{s\in C_i^+} \exp(\kappa \pmb{\mu}_s^\top \pmb{v}_i) }{ \sum_{l\neq c} \exp(\kappa \pmb{\mu}_l^\top \pmb{v}_i) }.
\end{align}
The total loss is the average over all pixels.
Minimizing this loss is to maximize the expected number of pixels correctly classified by associating the right neighbor prototypes.
The ground truth labels are thus used for finding the set of same-class segments $C^+_i$ w.r.t. pixel $i$ within a mini-batch (and memory banks).
If there is no other segment in the same category, we fall back to the previous vMF loss.
Since both vMF and vMF-N losses serve the same purpose for grouping and separating segments by optimizing the CNN feature extraction, we dub them segment sorting losses.

Understandably, an essential component of the segment sorting loss is the existence of semantic neighbor segments (in the numerator) and the abundance of alien segments (in the denominator).
That is, the more examples presented at once, the better the optimization.
We thus leverage two strategies:
1) We calculate the loss w.r.t. all the segments in the batch as opposed to traditionally image-wise loss function.
2) We use additional memory banks that cache the segment prototypes from previous batches.
In our experiments, we cache up to $2$ batches.
These two strategies help the fragmented segments (produced by segment partition in Fig.~\ref{fig:align}) connect to other similar segments between different images, or even between different batches.

%
%

\subsection{Inference via K-Nearest Neighbor Retrieval}
\label{sec:inference}

After training, we calculate and save all the segment prototypes in the training set.
We calculate the prototypes using pixels with majority labels within the segments, ignoring other unresolved noisy pixels.

During inference, we again conduct the K-Means clustering and then perform k-nearest neighbor search for each segment to retrieve the labels from segments in the training set.
The ablation study on inference runtime and memory can be found in the supplementary.

Our overall framework is non-parametric. We use vMF clustering to organize pixel embeddings into segment exemplars, whose number is proportional to number of images in the training set. The embeddings of exemplars are trained with a nearest neighbor criterion such that the inference can be done consistently, resulting in a non-parametric model.

\begin{table}[b]
  \centering
  \resizebox{0.8\linewidth}{!}{%
  \begin{tabular}{l|c|c}
    \Xhline{1pt}
    Base / Backbone / Method & mIoU & f-measure\\
    \hline \hline
    \rowcolor{Gray}
    DeepLabv3+ / MNV2 / Softmax  & 72.51 & 50.90 \\
    
    \rowcolor{Gray}
    DeepLabv3+ / MNV2 / SegSort & 74.94 & 58.83\\
    \hline
    
    \rowcolor{Gray}
    PSPNet / ResNet-101 / Softmax & 80.12 & 59.64\\
    
    \rowcolor{Gray}
    PSPNet / ResNet-101 / ASM~\cite{hwang2019adversarial} & 81.43 & 62.35 \\
    
    \rowcolor{Gray}
    PSPNet / ResNet-101 / SegSort & 81.77 & 63.71 \\
    \hline \hline
    
    DeepLabv3+ / MNV2 / Softmax & 73.25 & - \\
    
    DeepLabv3+ / MNV2 / SegSort & 74.88 & - \\
    \hline
    
    PSPNet / ResNet-101 / Softmax & 80.63 & -\\
    
    PSPNet / ResNet-101 / SegSort & 82.41 & - \\
    
    \Xhline{1pt}
    \end{tabular}}
    \vspace{0.5pt}
    \caption{Quantitative results on Pascal VOC 2012. The first 4 rows with gray colored background are on validation set while the last 4 rows are on testing set. Networks trained with SegSort consistently outperform their parametric counterpart (Softmax) by 1.63 to 2.43\% in mIoU and by 4.07 to 7.97\% in boundary f-measure.}
    \label{tab:voc}
\end{table}

\section{Experiments}
\label{sec:exp}

In this section, we demonstrate the efficacy of our Segment Sorting (SegSort) through experiments and visual analyses. We first describe the experimental setup in Section \ref{sec:exp_setup}. Then we summarize all the quantitative and qualitative results of fully supervised semantic segmentation in Section \ref{sec:exp_fully}. Lastly, we present results of the proposed approach for unsupervised semantic segmentation in Section \ref{sec:exp_unsupervised}. Additional experiments including ablation studies, t-SNE embedding visualization, and qualitative results on Cityscapes can be found in the supplementary.

\subsection{Experimental Setup}
\label{sec:exp_setup}

\noindent \textbf{Datasets.} We mainly use two datasets in the experiments, i.e., 
PASCAL VOC 2012 \cite{pascal-voc-2012} and Cityscapes \cite{cordts2016cityscapes}. 

PASCAL VOC 2012 \cite{pascal-voc-2012} segmentation dataset contains $20$ object categories and one background class. The original dataset contains $1,464$ ({\it train}) / $1,449$ ({\it val}) / $1,456$ ({\it test}) images.  Following the procedure of \cite{long2015fully, chen2016deeplab,zhao2016pyramid}, we augment the training data with the annotations of \cite{hariharan2011semantic}, resulting in $10,582$ ({\it train\_aug}) images. 

Cityscapes \cite{cordts2016cityscapes} is a dataset for semantic urban street scene understanding.  $5,000$ high quality pixel-level finely annotated images are divided into  training, validation, and testing sets with $2,975$ / $500$ / $1,525$ images, respectively. It defines $19$ categories containing flat, human, vehicle, construction, object, nature, etc. \\

\vspace{-6pt}
\noindent \textbf{Segmentation Architectures.} We use DeepLabv3+ \cite{deeplabv3plus2018} and PSPNet \cite{zhao2016pyramid} as the segmentation architectures, powered by MobileNetV2 \cite{mobilenetv22018} and ResNet101~\cite{he2016deep}, respectively, both of which are pre-trained on ImageNet \cite{krizhevsky2012imagenet}.

We follow closely the training procedures of the base architectures when training the baseline model with the standard pixel-wise softmax cross-entropy loss. The performance of the final model might be slightly worse from what is reported in the original papers mainly due to two reasons: 1) We do not pre-train on any other segmentation dataset, such as MS COCO \cite{lin2014microsoft} dataset. 2) We do not adopt any additional training tricks, such as balance sampling or fine-tuning specific categories. \\

\vspace{-6pt}
\noindent \textbf{Hyper-parameters of SegSort.}
For all the experiments, we use the following hyper-parameters for training SegSort: The dimension of embeddings is $32$. The number of clustering in K-Means are set to $25$ and $64$ for VOC and Cityscapes, respectively. The EM steps in K-Means are set to $10$ and $15$ for VOC and Cityscapes, respectively. The concentration constant is set to $10$. During inference, we use the same hyper-parameters for K-Means segmentation and $21$ nearest neighbors for predicting categories.

We use different learning rates and iterations for supervised training with SegSort. For VOC 2012, we train the network with initial learning rate $0.002$ for $100$k iterations on {\it train\_aug} set and with initial learning rate $0.0002$ for $30$k iterations on {\it train} set. For Cityscapes, we train the network with initial learning rate $0.005$ for the same 90k iterations as the softmax baseline.

Training on VOC 2012 requires more iterations than the baseline procedure mainly because most images only contain very few categories while the network can only compare segments in $3$ batches ($2$ batches were cached). We find that enlarging the batch size or increasing memory banks might reduce the training iterations. As a comparison, images from Cityscapes contain ample categories, so the training iterations remain the same.

\begin{table*}[t]
  \centering
  \resizebox{\textwidth}{!}{%
  \begin{tabular}{l|c c c c c c c c c c c c c c c c c c c|c}
    \Xhline{1pt}
    Method & road & swalk & build. & wall & fence & pole & tlight & tsign & veg. & terrain & sky & person & rider & car & truck & bus & train & mbike & bike & mIoU \\
    \hline \hline
    
    Softmax & 97.96 & 83.89 & 92.22 & 57.24 & 59.31 & 58.89 & 68.39 & 77.07 & 92.18 & 63.71 & 94.42 & 81.80 & 63.11 & 94.85 & 73.54 & 84.82 & 67.42 & 69.34 & 77.42 & 76.72 \\
    
    SegSort & 98.18 & 84.86 & 92.75 & 55.63 & 61.57 & 63.72 & 71.66 & 80.01 & 92.62 & 64.64 & 94.65 & 82.32 & 62.75 & 95.08 & 77.27 & 87.07 & 78.89 & 63.63 & 77.51 & 78.15 \\

    \Xhline{1pt}
    \end{tabular}}
    \vspace{0.5pt}
    \caption{Per-class results on Cityscapes validation set. We conclude that network trained with SegSort outperforms Softmax consistently.}
    \label{tab:cityscapes}
\end{table*}

\begin{figure}
    \centering
    \includegraphics[width=1.0\linewidth]{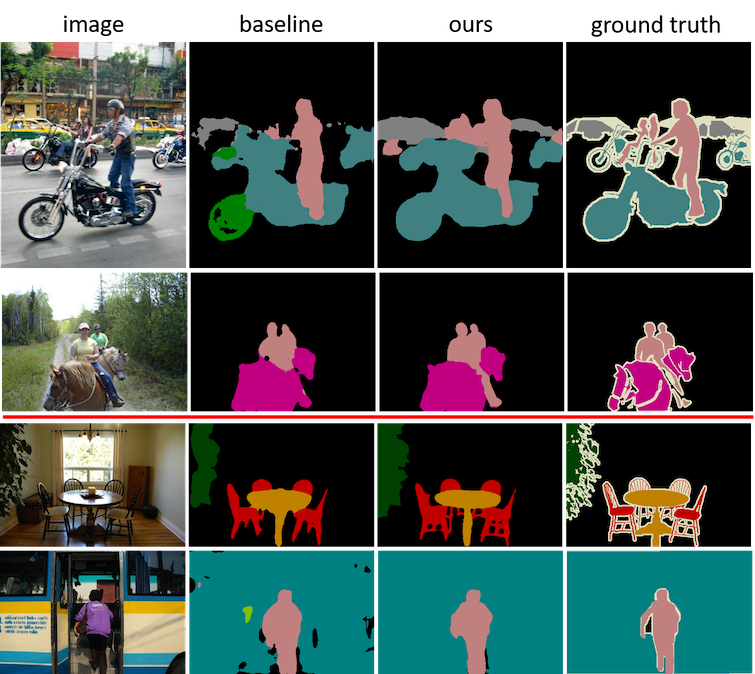}
    \caption{Visual comparison on PASCAL VOC 2012 validation set. We show the visual examples with DeepLabv3+ (upper 2 rows) and PSPNet (lower 2 rows). We observe prominent improvements on thin structures, such as human leg and chair legs. Also, more consistent region predictions can be observed when context is critical, such as wheels in motorcycles and big trunk of buses.}
    \label{fig:visual}
\end{figure}

\begin{figure}
    \centering
    \includegraphics[width=1.0\linewidth]{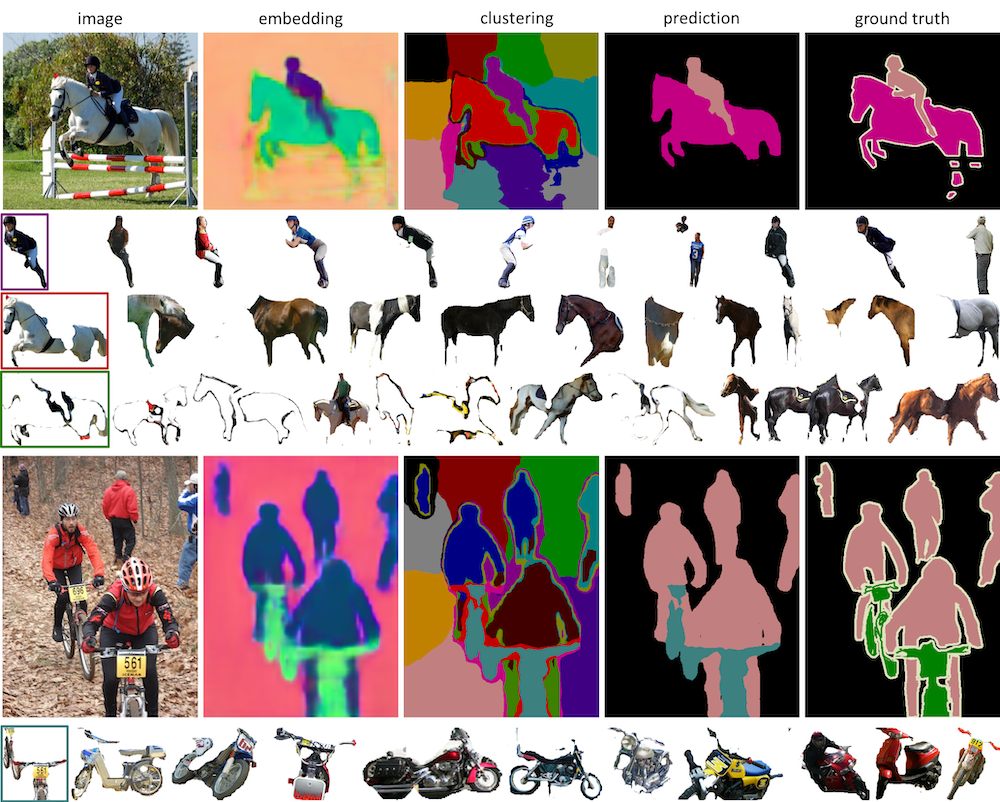}
    \caption{Two examples, correct and incorrect predictions, for segment retrieval for supervised semantic segmentation on VOC 2012 validation set. Query segments (leftmost) are framed by the same color in clustering. (Top) The query segments of rider, horse, and horse outlines can retrieve corresponding semantically relevant segments in the training set. (Bottom) For the failure case, it can be inferred from the retrieved segments that the number tag on the front of bikes is confused by the other number tags or front lights on motorbikes, resulting in false predictions.}
    \label{fig:retrieval}
\end{figure}

\subsection{Fully Supervised Semantic Segmentation}
\label{sec:exp_fully}

\noindent \textbf{VOC 2012:} We summarize the quantitative results of fully supervised semantic segmentation on Pascal VOC 2012 \cite{pascal-voc-2012} in Table \ref{tab:voc}, evaluated using mIoU and boundary evaluation following~\cite{arbelaez2011contour,aaf2018} on both validation and testing set.

We conclude that networks trained with SegSort consistently outperform their parametric counterpart (Softmax) by $1.63$ to $2.43\%$ in mIoU and by $4.07$ to $7.97\%$ in mean boundary f-measure. (Per-class results can be found in the supplementary.) We notice that SegSort with DeepLabv3+ / MNV2 captures better fine structures, such as in `bike' and `mbike' while with PSPNet / ResNet-101 enhances more towards detecting small objects, such as in `boat' and `plant'.

We present the visual comparison in Fig. \ref{fig:visual}. We observe prominent improvements on thin structures, such as human legs and chair legs. Also, more consistent region predictions can be found when context is critical, such as wheels in motorcycles and big trunk of buses.

One of the most important features of SegSort is the self-explanatory predictions via nearest neighbor segment retrieval. We therefore demonstrate two examples, correct and incorrect predictions, in Fig. \ref{fig:retrieval}. As can be seen, the query segments (on the leftmost) of rider, horse, and horse outlines can retrieve corresponding semantically relevant segments in the training set. For the incorrect example, it can be inferred from the retrieved segments that the number tag on the front of bikes was confused by the other number tags on motorbikes, resulting in false predictions. \\

\vspace{-6pt}
\noindent \textbf{Cityscapes:} We summarize the quantitative results of fully supervised semantic segmentation on Cityscapes \cite{cordts2016cityscapes} in Table \ref{tab:cityscapes}, evaluated on the validation set. Due to limited space, visual results are included in the supplementary. 

The network trained with SegSort outperforms Softmax consistently. Large objects, \eg, `bus' and `truck', are improved thanks to more consistent region predictions while small objects, \eg, `pole' and `tlight', are better captured.

\begin{figure}
    \centering
    \includegraphics[width=1.0\linewidth]{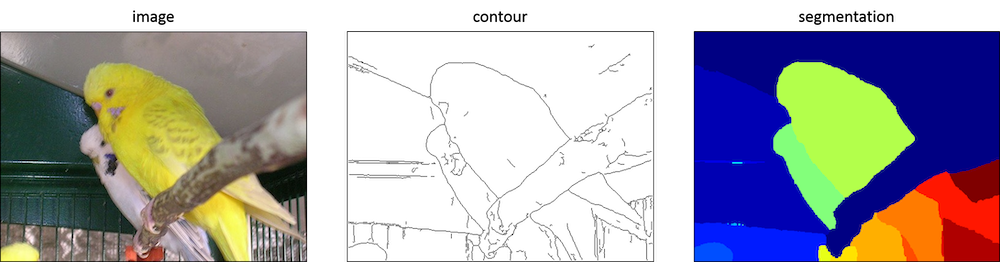}
    \caption{Training data for unsupervised semantic segmentation. We produce fine segmentations (right), HED-owt-ucm, from the contours (middle) detected by HED \cite{xie2015holistically}, followed by the procedure in gPb-owt-ucm \cite{malik2001contour}. }
    \label{fig:unsup_data}
\end{figure}

\begin{table}
  \centering
  \resizebox{1.0\linewidth}{!}{%
  \begin{tabular}{| c | c | c ||c| c|}
    \Xhline{1pt}
    unsup. on \textit{train\_aug} & sup. on \textit{train\_aug} & sup. on \textit{train} & mIoU & f-measure \\
    \hline \hline
    
     &  &  & 49.50 & 40.86 \\
    \hline
    
    \checkmark &  &  & 55.86 & 44.78 \\
    \hline
    
    & & \checkmark & 71.86  & 55.70 \\
    \hline
    
    \checkmark & & \checkmark & 72.47 & 56.52 \\
    \hline
    
    & \checkmark & \checkmark & 73.35 & 55.69 \\
    
    \Xhline{1pt}
    \end{tabular}
    }
    \vspace{0.5pt}
    \caption{Quantitative results related to unsupervised semantic segmentation on Pascal VOC 2012 validation set. Our unsupervised trained network (2$^\text{nd}$ row) outperforms the baseline (1$^\text{st}$ row) of directly clustering pretrained features using HED-owt-ucm~\cite{xie2015holistically} and achieves $76\%$ performance of its supervised counterpart (5$^\text{th}$ row). Also, the network fine-tuned from unsupervised pre-trained embeddings (4$^\text{th}$ row) outperforms the one without (3$^\text{rd}$ row) in both mIoU and boundary f-measure.}
    \label{tab:unsupervised}
\end{table}

\begin{figure}
    \centering
    \includegraphics[width=1.0\linewidth]{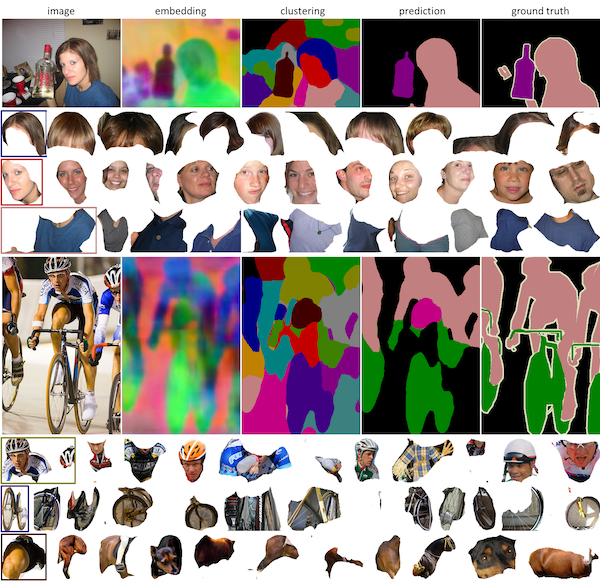}
    \caption{Segment retrieval results for unsupervised semantic segmentation on VOC 2012 vadlidation set. Query segments (leftmost) are framed by the same color in clustering. As is observed, the embeddings learned by unsupervised SegSort attend to more visual than semantic similarities compared to the supervised setting. Hairs, faces, blue shirts, and wheels are retrieved successfully. The last query segment fails because the texture around knee is more similar to animal skins.}
    \label{fig:unsup_retrieval}
    \vspace{-6pt}
\end{figure}

\subsection{Unsupervised Semantic Segmentation}
\label{sec:exp_unsupervised}

We train the model using our framework {\bf without} any ground truth labels at any level, pixel-wise or image-wise. 

To adapt our approach for unsupervised semantic segmentation, what we need is a good criterion for segmenting an image along visual boundaries, which serves as a pseudo ground truth mask. There is an array of methods that meet the requirement, \eg, SLIC \cite{achanta2012slic} for super-pixels or gPb-owt-ucm \cite{arbelaez2011contour} for hierarchical segmentation. We choose the HED contour detector \cite{xie2015holistically} pretrained on BSDS500 dataset \cite{arbelaez2011contour}, and follow the procedure in gPb-owt-ucm \cite{arbelaez2011contour} to produce the hierarchical segmentation, or HED-owt-ucm (Fig. \ref{fig:unsup_data}).

We train the PSPNet / ResNet-101 network on the same augmented training set on VOC 2012 as in the supervised setting with the same initial learning rate yet for only $10$k iterations. The hyper-parameters remain unchanged.

Note that the contour detector only provides visual boundaries without any concept of semantic segments, yet through our feature learning with segment sorting, our method discovers segments of common features -- {\it semantic segmentation without names}.

For the sake of performance evaluation, we assume there is a separate annotated image set available during inference. For each segment under query, we assign a label by  the  majority  vote  of  its  nearest  neighbors  from that annotated set. 

Table \ref{tab:unsupervised} shows that our unsupervised trained network outperforms the baseline of directly clustering pretrained features using HED-owt-ucm~\cite{xie2015holistically} segmentation and further achieves $76\%$ performance of its supervised counterpart. Together, We also showcase one possible way to make use of the unsupervised learned embedding. The network fine-tuned from unsupervised pre-trained embeddings outperforms the one without. 
Fig. \ref{fig:unsup_retrieval} shows the embeddings learned by unsupervised SegSort attend to more visual than semantic similarities compared to the supervised setting because the fine segmentation formed by contour detectors partitions the image into visually consistent segments. Hairs, faces, blue shirts, and wheels are retrieved successfully. The last query segment fails because the texture around the knee is more similar to animal skins. \\

\begin{figure}
    \centering
    \includegraphics[width=1.0\linewidth]{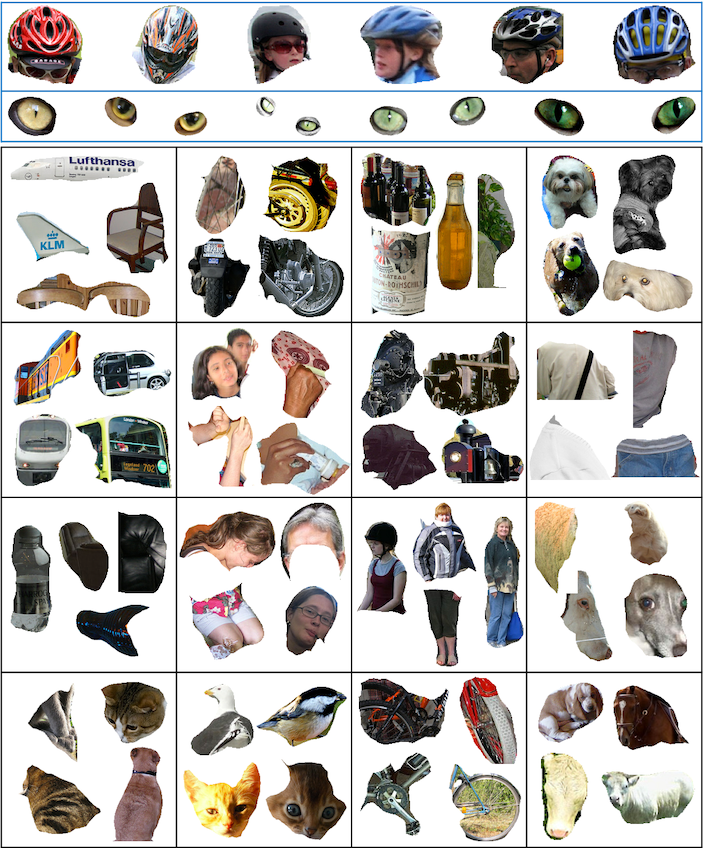}
    \caption{We perform a nearest neighbor based hierarchical agglomerative clustering, FINCH~\cite{sarfraz2019efficient} on foreground segment prototypes to discover visual groups. Top two rows show random samples from two clusters at the finest level. Bottom table displays clusters at a coarser level of $16$ clusters. We show four representative segments per cluster. }
    \label{fig:finch}
    \vspace{-6pt}
\end{figure}

\vspace{-6pt}
\noindent {\bf Automatic Discovery of Visual Groups.}
We noticed in the retrieval results that CNNs trained this way can discover visual groups.
We wonder if such visual structures actually form different clusters (or fine categories).

We extract all foreground segments in the training set and perform a nearest neighbor based hierarchical agglomerative clustering algorithm FINCH~\cite{sarfraz2019efficient}.
FINCH merges two points if one is the nearest neighbor of the other (with undirectional link).
This procedure can be performed recursively.
We start with $1501$ segment prototypes and performs FINCH to produce $252$, $57$, $16$, $3$, and $1$ clusters after each iteration.
We visualize some segment groups at the finest level and a coarser level of $16$ clusters in Fig.~\ref{fig:finch}.

A bigger picture of how the segments relate to each other from t-SNE~\cite{maaten2008visualizing} can be found in the supplementary.

\section{Conclusion}
\label{sec:con}

We proposed an end-to-end pixel-wise metric learning approach that is motivated by perceptual organization.
We integrated the two essential components, pixel-level and segment-level sorting, in a unified framework, derived from von Mises-Fisher clustering.
We demonstrated the proposed approach consistently improves over the conventional pixel-wise prediction approaches for supervised semantic segmentation.
We also presented the first attempt for unsupervised semantic segmentation.
Intriguingly, the predictions produced by our approach, correct or not, can be inherently explained by the retrieved nearest segments. \\

\vspace{-6 pt}
\noindent {\bf Acknowledgements.} This research was supported, in part, by Berkeley Deep Drive, NSF (IIS-1651389), DARPA.

\newpage

{\small
\bibliographystyle{ieee_fullname}
\bibliography{egbib}
}

\clearpage


\section{Supplementary}

\subsection{Algorithm}
\label{sec:alg}

We summarize the overall algorithm, trained with supervision, in Algorithm~\ref{alg}. The unsupervised algorithm uses the oversegmentation from object detector as pixel sorting and Equation~\ref{eqn:vmf_loss} for loss calculation, which encourages pixel embeddings in each segment to form an isolated cluster.

\begin{algorithm}[b]
\label{alg}
\SetAlgoLined
\SetNoFillComment
\For{number of training iterations} {
    Sample a minibatch with $m$ images $\{ \pmb{x}^{(1)}, \dots, \pmb{x}^{(m)} \}$ and segmentation masks $\{ \pmb{y}^{(1)}, \dots, \pmb{y}^{(m)} \}$. \\
    Extract the deep feature embedding $\{ \pmb{v}^{(1)}, \dots, \pmb{v}^{(m)} \}$ for each image.
    Compute $\pmb{v}'$ by concatenating $\pmb{v}$ with coordinate features. \\
    Initialize $\pmb{R}$ (segment IDs) by uniformly partitioning each image for $k$ regions.
    
    \tcc{K-Means clustering.}
    \For{number of K-Means iterations} {
    \tcc{The M step.} 
    \For{each region $j$}{
    $\pmb{\mu}'_j \leftarrow \sum_{k \in \pmb{R}_j} \pmb{v}'_k / ||\sum_{k \in \pmb{R}_j} \pmb{v}'_k||$ 
    }
    
    \tcc{The E step.}
    $\pmb{R} \leftarrow \argmax \pmb{\mu}'^\top \pmb{v}'$
    }
    
    Partition a segment if it contains multiple labels.
    
    \tcc{Compute prototypes.}
    \For{each region $j$}{
    $\pmb{\mu}_j \leftarrow \sum_{k \in \pmb{R}_j} \pmb{v}_k / ||\sum_{k \in \pmb{R}_j} \pmb{v}_k||$ 
    }
    
    \tcc{Calculate vMF-N loss.}
    Calculate the loss using Equation 10 and back-propagate the errors.
    }
\caption{Supervised SegSort algorithm.}
\end{algorithm}

\subsection{t-SNE Embedding Visualization}
\label{sec:tsne}

\begin{figure*}
    \centering
    \includegraphics[height=1.0\textheight]{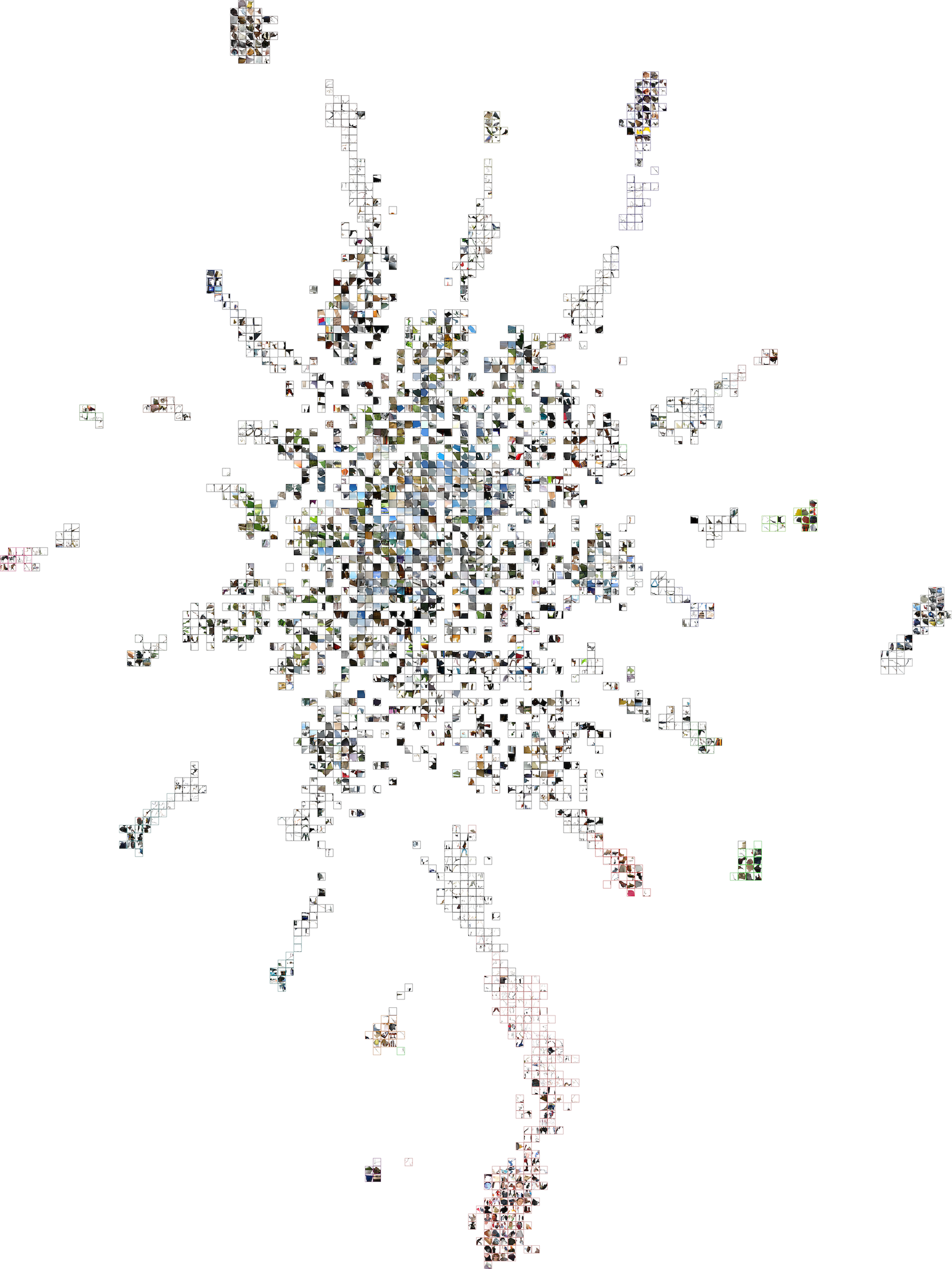}
    \caption{t-SNE visualization of prototype embeddings from {\bf supervised} SegSort, framed with category color. Best viewed with zoom-in.}
    \label{fig:sup_tsne_visualization}
\end{figure*}

\begin{figure*}
    \centering
    \includegraphics[height=1.0\textheight]{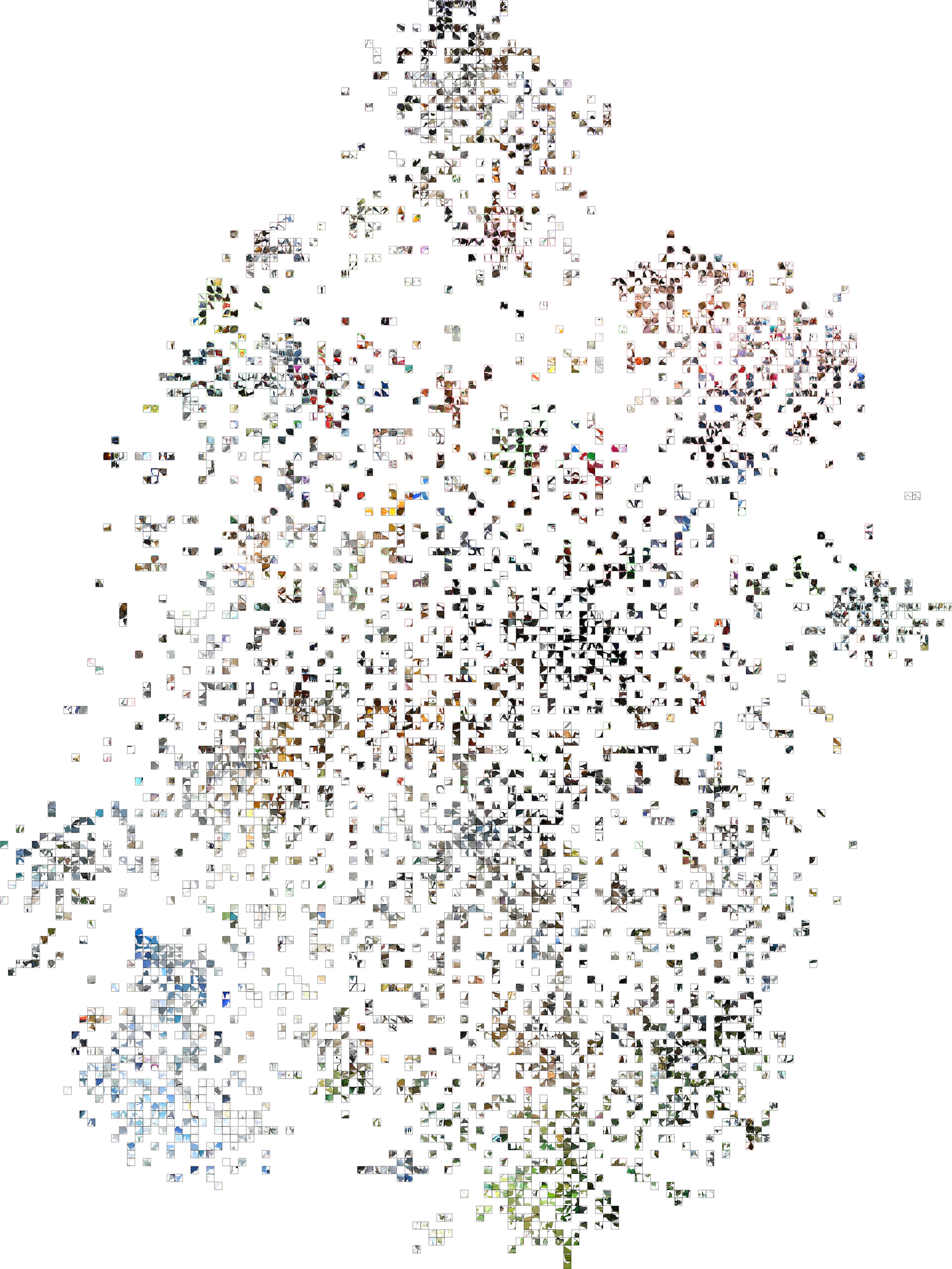}
    \caption{t-SNE visualization of prototype embeddings from {\bf unsupervised} SegSort, framed with category color. Best viewed with zoom-in.}
    \label{fig:unsup_tsne_visualization}
\end{figure*}

We visualize the prototype embeddings in the training set using t-SNE \cite{maaten2008visualizing}. We display the results from {\bf supervised} and {\bf unsupervised} SegSort in Figure \ref{fig:sup_tsne_visualization} and \ref{fig:unsup_tsne_visualization}, respectively. This is done by random sampling $5,000$ prototypes in the training set, reducing the dimension from $32$ to $2$ using t-SNE \cite{maaten2008visualizing}, and placing the corresponding patches on the 2D canvas wherever possible. 

For visualization of supervised SegSort, we observe that most background patches form a large cluster in the center with some small visual clusters. Each stretching arm represents one foreground class, with gradual appearance changes from boundaries to object centers. For examples, cars and trucks are on the rightmost islands while horses, cows, and sheeps on the leftmost.

For visualization of unsupervised SegSort, we observe that clusters are formed more by visual similarities. The cues for clustering are usually color and texture. For examples, wheels are clustered on the rightmost island while animals on the top. Grass and sky are placed on the bottom.

\subsection{Study on Inference Latency and Memory}

We analyze, compared to Softmax baseline, SegSort's inference latency and memory as they are of practical concern. We conclude the runtime overhead ($7\%$-$22\%$) is manageable and memory overhead ($\sim1.5\%$) is negligible.

We conduct experiments with various k-means iterations and numbers of nearest neighbors to learn how they influence the inference performance, summarized in Table~\ref{tab:inference_ablation}. All experiments (PSPNet inference at single scale) are done using the same GTX 1080 Ti GPU. The overall GPU memory usage overhead is only $\sim$1.5\% as no extra parameters are introduced. The runtime overhead is $7\%$-$22\%$. We also notice the most runtime overhead is due to k-means instead of kNN (with 36K prototypes), both of which are computed in GPU. With 4 k-means iterations and 11-NN, our method (with $11\%$ runtime overhead) already improves more than $1.5\%$ mIoU. We believe this  latency/accuracy trade-off is reasonable, particularly with the benefits such as interpretability.

\begin{table}[h]
  \vspace{-8pt}
  \centering
  \resizebox{0.48\textwidth}{!}{%
  \begin{tabular}{|l||c | c | c | c | c | }
    \Xhline{1pt}
    Method & k-means \% time & kNN \% time & Overall time & memory & mIoU \\
    \hline \hline
    
    Softmax &  - & - & 170.34 & 4504 & 76.96 \\
    
    2 iter 1-NN  & 11.43 & 4.78 & 182.19 & 4569 & 77.18 \\
    
    4 iter 1-NN & 13.97 & 4.60 & 188.61 & 4573 & 78.05 \\
    
    4 iter 11-NN & 13.72 & 4.67 & 189.09 & 4574 & 78.50 \\
    
    10 iter 11-NN & 20.04 & 4.38 & 207.51 & 4577 & 78.69 \\

    \Xhline{1pt}
    \end{tabular}}
    \caption{Ablation study on runtime (ms) and GPU memory (MiB).}
    \vspace{-9pt}
    \label{tab:inference_ablation}
\end{table}

\subsection{Boundary Evaluation}

We explain how we conduct boundary evaluation on semantic segmentation following~\cite{arbelaez2011contour, aaf2018}. We first compute semantic boundaries per category for the semantic predictions and ground truth. We then match boundary pixels between predictions and ground truth with maximum distance of $0.01$ of image diagonal length. The per-category results are summarized by precision, recall, and f-measure in Table~\ref{tab:voc_boundary} and Table~\ref{tab:cityscapes_boundary} on VOC and Cityscapes datasets.

\begin{table*}
  \centering
  \resizebox{\textwidth}{!}{%
  \begin{tabular}{l|c c c c c c c c c c c c c c c c c c c c|c}
    \Xhline{1pt}
    Base / Method & aero & bike & bird & boat & bottle & bus & car & cat & chair & cow & table & dog & horse & mbike & person & plant & sheep & sofa & train & tv & mean \\
    \hline \hline

    Deeplabv3$^+$ / Softmax & 76.88 & 69.30 & 70.41 & 52.87 & 44.73 & 61.37 & 62.44 & 65.08 & 38.14 & 68.08 & 20.58 & 54.64 & 63.41 & 60.84 & 62.48 & 49.75 & 68.12 & 30.40 & 51.05 & 44.53 &  56.02 \\

    Deeplabv3$^+$ / SegSort & 81.70 & 72.53 & 75.13 & 62.07 & 69.34 & 72.42 & 66.67 & 76.10 & 45.23 & 70.91 & 40.72 & 69.55 & 65.09 & 72.75 & 74.24 & 53.70 & 80.07 & 43.70 & 70.86 & 61.98 & 66.61 \\
    \hline
    
    PSPNet / Softmax & 91.10 & 75.72 & 90.72 & 68.00 & 68.19 & 82.00 & 73.49 & 81.58 & 54.96 & 85.38 & 40.57 & 76.17 & 84.00 & 77.07 & 76.55 & 65.17 & 87.65 & 44.63 & 73.53 & 61.95 & 72.97 \\

    PSPNet / SegSort & 86.46 & 73.81 & 84.86 & 68.65 & 74.18 & 81.51 & 74.48 & 81.43 & 57.92 & 83.24 & 52.58 & 75.71 & 81.29 & 74.64 & 79.17 & 63.95 & 86.80 & 43.57 & 68.01 & 63.45 & 73.05 \\
    \hline
    \hline
    
    Deeplabv3$^+$ / Softmax & 64.03 & 41.80 & 60.19 & 36.93 & 49.53 & 57.85 & 52.31 & 60.25 & 26.10 & 54.54 & 21.65 & 59.90 & 55.82 & 47.89 & 49.37 & 21.90 & 46.70 & 38.14 & 53.02 & 41.34 & 47.52 \\

    Deeplabv3$^+$ / SegSort & 70.25 & 47.19 & 61.98 & 42.03 & 58.34 & 61.81 & 54.95 & 64.93 & 33.87 & 54.85 & 30.43 & 63.99 & 59.40 & 54.41 & 56.76 & 26.61 & 48.35 & 46.85 & 54.70 & 57.59 & 53.20 \\
    \hline
    
    PSPNet / Softmax & 68.07 & 37.66 & 63.66 & 38.90 & 59.55 & 62.05 & 55.28 & 64.92 & 29.48 & 55.87 & 28.16 & 65.18 & 58.09 & 49.35 & 53.69 & 22.75 & 52.09 & 41.14 & 55.74 & 51.78 &  51.19 \\

    PSPNet / SegSort & 70.97 & 38.66 & 70.07 & 46.51 & 67.99 & 65.11 & 61.48 & 68.46 & 40.73 & 61.71 & 39.15 & 69.96 & 63.22 & 54.04 & 59.59 & 30.64 & 53.69 & 52.46 & 58.04 & 61.14 & 57.29 \\
    \hline
    \hline
    
    Deeplabv3$^+$ / Softmax & 69.87 & 52.14 & 64.90 & 43.48 & 47.01 & 59.56 & 56.93 & 62.57 & 30.99 & 60.56 & 21.10 & 57.15 & 59.37 & 53.60 & 55.16 & 30.41 & 55.41 & 33.83 & 52.01 & 42.88 & 50.90 \\

    Deeplabv3$^+$ / SegSort & 75.55 & 57.18 & 67.92 & 50.12 & 63.37 & 66.70 & 60.25 & 70.07 & 38.74 & 61.86 & 34.83 & 66.65 & 62.11 & 62.26 & 64.33 & 35.59 & 60.29 & 45.22 & 61.74 & 59.71 & 58.83 \\
    \hline
    
    PSPNet / Softmax & 77.92 & 50.30 & 74.82 & 49.49 & 63.58 & 70.64 & 63.10 & 72.30 & 38.38 & 67.55 & 33.24 & 70.25 & 68.68 & 60.17 & 63.11 & 33.73 & 65.34 & 42.82 & 63.41 & 56.41 & 59.64 \\

    PSPNet / SegSort & 77.95 & 50.75 & 76.76 & 55.45 & 70.95 & 72.39 & 67.36 & 74.38 & 47.83 & 70.88 & 44.88 & 72.73 & 71.13 & 62.69 & 68.00 & 41.43 & 66.34 & 47.61 & 62.63 & 62.27 & 63.71 \\

    \Xhline{1pt}
    \end{tabular}}
    \vspace{0.5pt}
    \caption{Per-class boundary evaluation on Pascal VOC 2012 validation set. From top to bottom: precision, recall, and f-measure, separated by double lines.}
    \label{tab:voc_boundary}
\end{table*}

\begin{table*}[t]
  \centering
  \resizebox{\textwidth}{!}{%
  \begin{tabular}{l|c c c c c c c c c c c c c c c c c c c|c}
    \Xhline{1pt}
    Method & road & swalk & build. & wall & fence & pole & tlight & tsign & veg. & terrain & sky & person & rider & car & truck & bus & train & mbike & bike & mean \\
    \hline \hline
    
    Softmax & 85.68 & 77.34 & 85.60 & 55.33 & 49.44 & 90.34 & 82.71 & 77.77 & 92.95 & 66.23 & 95.59 & 87.50 & 89.12 & 92.73 & 54.55 & 76.99 & 72.29 & 67.87 & 85.35 &  78.18 \\
    
    SegSort & 89.14 & 78.81 & 88.79 & 55.18 & 55.45 & 92.53 & 87.70 & 82.32 & 94.25 & 70.51 & 96.07 & 90.73 & 87.89 & 95.61 & 55.19 & 77.90 & 64.79 & 62.47 & 86.00 &  79.54 \\
    \hline \hline
    
    Softmax & 45.81 & 77.13 & 58.38 & 47.94 & 53.65 & 58.65 & 64.86 & 68.67 & 65.04 & 58.90 & 58.66 & 73.31 & 63.08 & 81.21 & 57.69 & 73.35 & 56.95 & 54.96 & 69.71 &  62.52 \\
    
    SegSort & 46.08 & 79.05 & 60.75 & 49.84 & 57.68 & 63.37 & 74.57 & 75.18 & 66.28 & 61.17 & 60.36 & 76.53 & 71.78 & 83.21 & 64.58 & 78.67 & 65.28 & 61.61 & 73.48 &  66.82 \\
    \hline \hline
    
    Softmax & 59.70 & 77.24 & 69.41 & 51.37 & 51.46 & 71.13 & 72.71 & 72.94 & 76.53 & 62.35 & 72.71 & 79.78 & 73.87 & 86.59 & 56.08 & 75.13 & 63.71 & 60.73 & 76.74 &  68.96 \\
    
    SegSort & 60.76 & 78.93 & 72.14 & 52.37 & 56.54 & 75.22 & 80.60 & 78.59 & 77.83 & 65.51 & 74.14 & 83.03 & 79.02 & 88.98 & 59.52 & 78.29 & 65.04 & 62.04 & 79.25 &  71.99 \\

    \Xhline{1pt}
    \end{tabular}}
    \vspace{0.5pt}
    \caption{Per-class boundary evaluation on Cityscapes validation set with PSPNet architecture. From top to bottom: precision, recall, and f-measure, separated by double lines.}
    \label{tab:cityscapes_boundary}
\end{table*}

\subsection{Ablation Study}
\label{sec:abl}
We conduct experiments for the ablation study to understand how different components affect the performance of supervised SegSort. We decide the hyper-parameters of our main experiments using the experiences learned from the ablation studies.

\begin{figure}
    \centering
    \includegraphics[width=\linewidth]{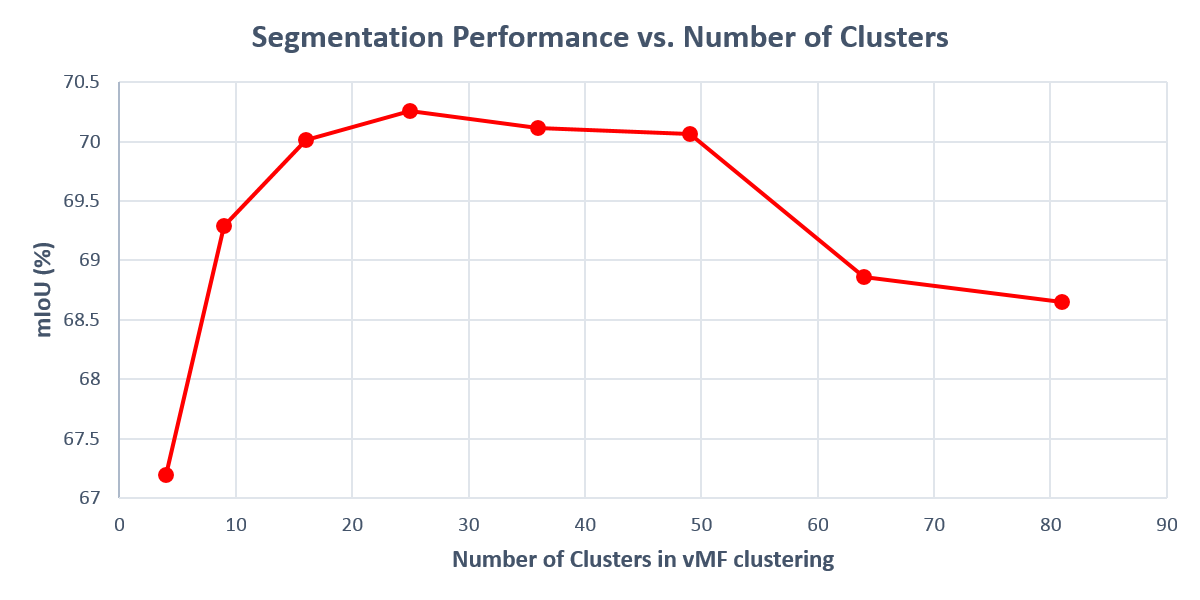}
    \caption{We show how the number of clusters affects the segmentation performance. The highest performance is at $25$ clusters, which are slightly more than the number of categories in the dataset.}
    \label{fig:abl_nc}
\end{figure}

{\bf Number of clusters:} We study how the number of clusters affects the semantic segmentation performance (Figure \ref{fig:abl_nc}). We train and test the DeepLabv3+ / MNV2 network with $4$, $9$, $16$, $25$, $36$, $49$, $64$, and $81$ clusters in the vMF clustering. The highest performance is at $25$ clusters, which are slightly more than the number of categories in the dataset.

\begin{figure}
    \centering
    \includegraphics[width=\linewidth]{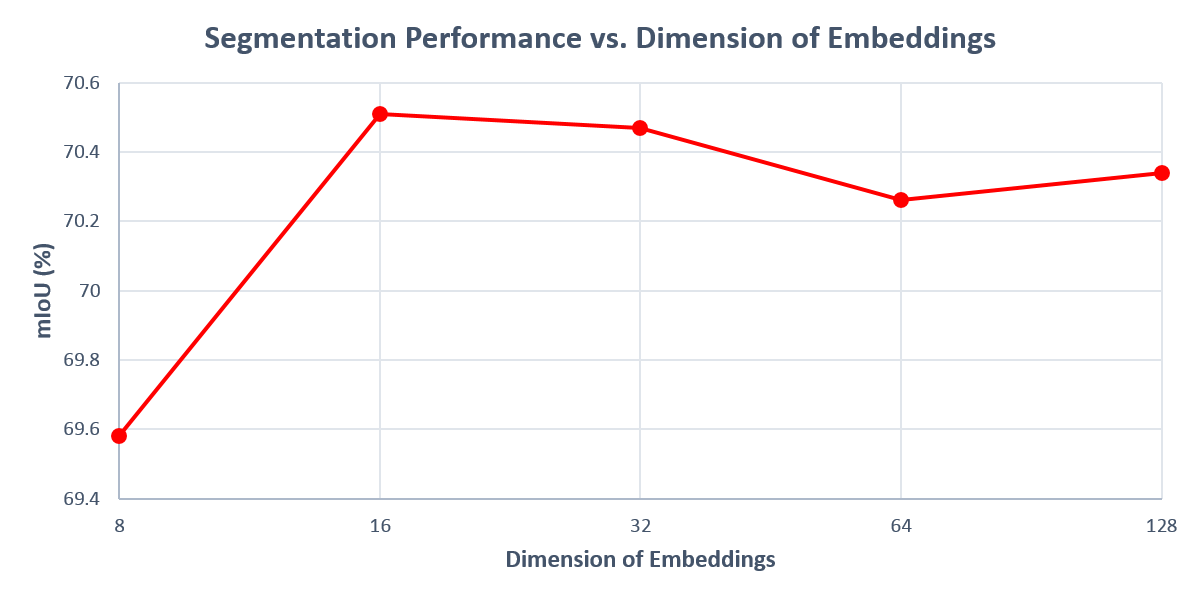}
    \caption{We study how the dimension of embeddings affects the segmentation performance. We conclude that as long as the embedding dimension is sufficient, \ie, larger than 8, the performance does not change drastically.}
    \label{fig:abl_dim}
\end{figure}

{\bf Dimension of embeddings:} We study how the dimension of embeddings affects the semantic segmentation performance (Figure \ref{fig:abl_dim}). We train and test the DeepLabv3+ / MNV2 network with $8$, $16$, $32$, $64$, and $128$ embedding dimension. We conclude that as long as the embedding dimension is sufficient, \ie, larger than 8, the performance does not change drastically.

\begin{figure}
    \centering
    \includegraphics[width=\linewidth]{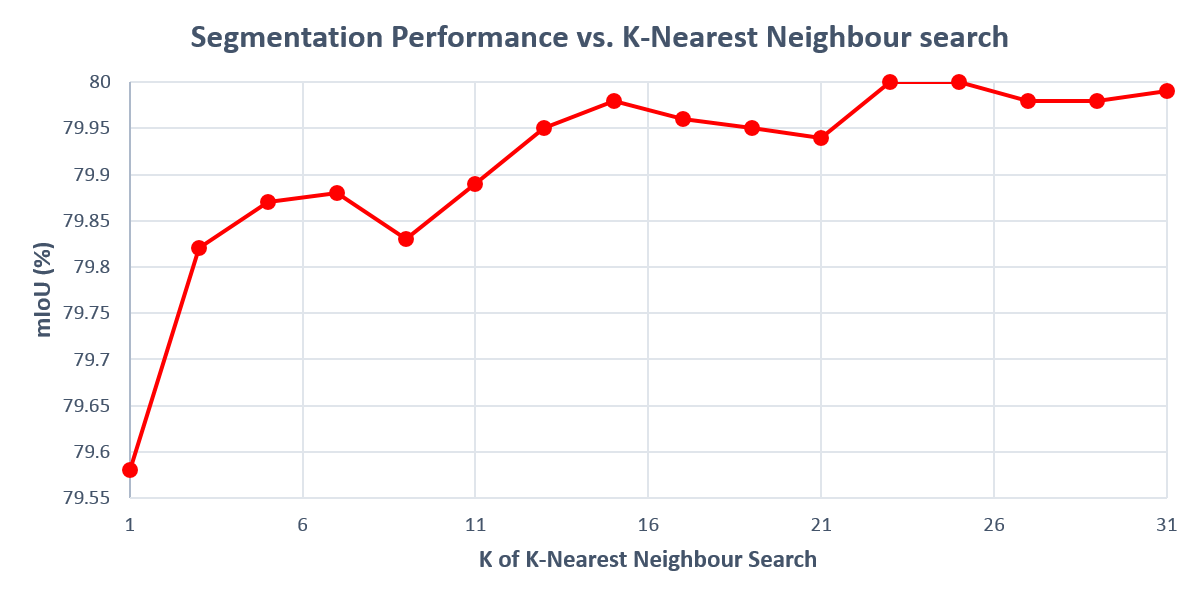}
    \caption{We study how the number of nearest neighbors during inference affects the segmentation performance. We conclude that the segmentation performance is robust to the number of nearest neighbors as the mIoU spans only $0.4\%$.}
    \label{fig:abl_knn}
\end{figure}

{\bf Number of nearest neighbors:} We study how the number of nearest neighbors during inference affects the segmentation performance  (Figure \ref{fig:abl_knn}). We train the PSPNet / ResNet-101 network as described in the main paper and test it using $1$ to $31$ (odd numbered) nearest neighbors. We conclude that the segmentation performance is robust to the number of nearest neighbors as the mIoU spans only $0.4\%$.

\subsection{Visualization and Test Results on Cityscapes}

\begin{figure*}
    \centering
    \includegraphics[width=0.9\linewidth]{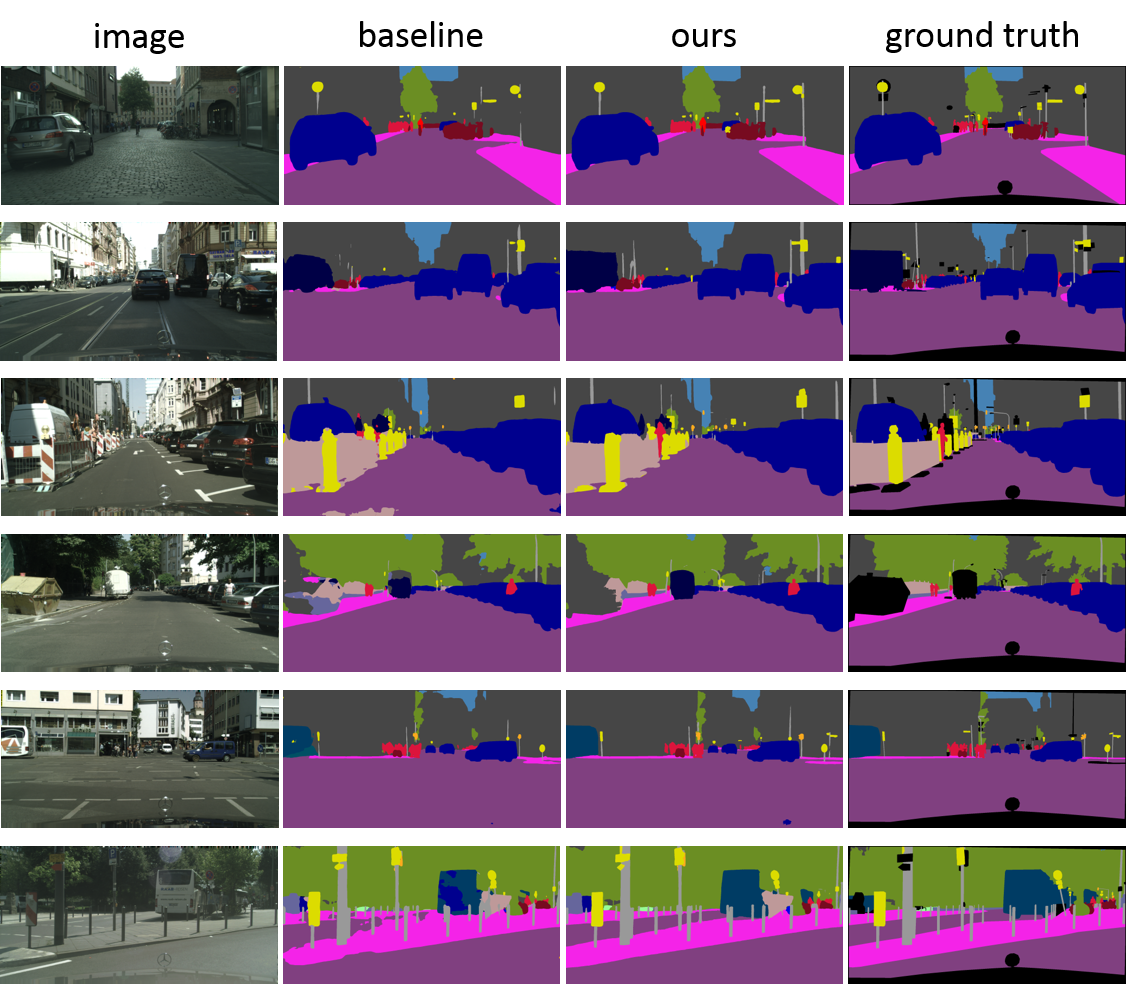}
    \caption{Visual comparison on Cityscapes validation set. We observe large objects, such as `bus' and `truck', are improved thanks to more consistent region predictions while small objects, such as `pole' and `tlight', are also better captured.}
    \label{fig:cityscapes_visual}
\end{figure*}

We present the visual comparison in Figure \ref{fig:cityscapes_visual}. We observe large objects, such as `bus' and `truck', are improved thanks to more consistent region predictions while small objects, such as `pole' and `tlight', are also better captured.

We also include the per-category segmentation performance on Cityscapes test set in Table \ref{tab:cityscapes_test}. We observe similar performance trends as on the validation set. We conclude that network trained with SegSort outperforms Softmax consistently.

\begin{table*}
  \centering
  \resizebox{\textwidth}{!}{%
  \begin{tabular}{l|c c c c c c c c c c c c c c c c c c c|c}
    \Xhline{1pt}
    Method & road & swalk & build. & wall & fence & pole & tlight & tsign & veg. & terrain & sky & person & rider & car & truck & bus & train & mbike & bike & mIoU \\
    \hline \hline
    
    Softmax & 98.33 & 84.21 & 92.14 & 49.67 & 55.81 & 57.62 & 69.01 & 74.17 & 92.70 & 70.86 & 95.08 & 84.21 & 66.58 & 95.28 & 73.52 & 80.59 & 70.54 & 65.54 & 73.73 & 76.30 \\
    
    SegSort & 98.56 & 85.84 & 92.85 & 52.18 & 58.44 & 63.62 & 73.29 & 77.93 & 93.41 & 72.37 & 95.10 & 84.97 & 67.98 & 95.41 & 67.43 & 80.80 & 60.69 & 68.68 & 74.34 & 77.05 \\

    \Xhline{1pt}
    \end{tabular}}
    \vspace{0.5pt}
    \caption{Per-class results on Cityscapes test set. We conclude that network trained with SegSort outperforms Softmax consistently.}
    \label{tab:cityscapes_test}
\end{table*}

\section{Per-category results on VOC}

We present the per-category results in Table~\ref{tab:voc_per_class} for interested readers. We notice that SegSort with DeepLabv3+ / MNV2 captures better fine structures, such as in ‘bike’ and ‘mbike’ while with PSPNet / ResNet-101 enhances more towards detecting small objects, such as in ‘boat’ and ‘plant’.

\begin{table*}
  \centering
  \resizebox{\textwidth}{!}{%
  \begin{tabular}{l|c c c c c c c c c c c c c c c c c c c c|c}
    \Xhline{1pt}
    Base / Backbone / Method & aero & bike & bird & boat & bottle & bus & car & cat & chair & cow & table & dog & horse & mbike & person & plant & sheep & sofa & train & tv & mIoU \\
    \hline \hline
    \rowcolor{Gray}
    DeepLabv3+ / MNV2 / Softmax & 85.02 & 55.18 & 80.92 & 65.87 & 70.60 & 89.55 & 83.39 & 88.27 & 35.04 & 80.30 & 48.24 & 79.20 & 82.13 & 81.16 & 81.21 & 52.59 & 75.24 & 47.20 & 80.20 & 67.92 &     72.51 \\
    
    \rowcolor{Gray}
    DeepLabv3+ / MNV2 / SegSort & 84.80 & 58.54 & 81.08 & 68.92 & 79.15 & 89.75 & 85.24 & 89.64 & 34.88 & 74.60 & 58.62 & 84.34 & 79.07 & 84.94 & 85.92 & 54.65 & 76.76 & 50.74 & 82.95 & 74.57 &    74.94 \\
    \hline
    
    \rowcolor{Gray}
    PSPNet / ResNet-101 / Softmax & 92.56 & 66.70 & 91.10 & 76.52 & 80.88 & 94.43 & 88.49 & 93.14 & 38.87 & 89.33 & 62.77 & 86.44 & 89.72 & 88.36 & 87.48 & 56.95 & 91.77 & 46.23 & 88.59 & 77.14 &     80.12 \\
    
    \rowcolor{Gray}
    PSPNet / ResNet-101 / SegSort & 92.23 & 52.68 & 91.29 & 80.33 & 83.92 & 95.13 & 90.33 & 95.44 & 44.68 & 90.84 & 67.37 & 91.29 & 91.09 & 89.66 & 88.98 & 67.54 & 88.06 & 53.04 & 87.79 & 79.97 &     81.77 \\
    \hline \hline
    
    DeepLabv3+ / MNV2 / Softmax & 85.89 & 59.20 & 79.09 & 61.24 & 66.47 & 87.87 & 85.17 & 88.80 & 28.27 & 78.98 & 60.67 & 80.35 & 83.72 & 83.90 & 83.52 & 59.87 & 83.43 & 50.22 & 74.07 & 63.91 &    73.25 \\
    
    DeepLabv3+ / MNV2 / SegSort & 79.49 & 66.32 & 75.38 & 66.17 & 70.71 & 91.51 & 84.82 & 85.54 & 38.69 & 74.91 & 68.99 & 78.17 & 80.49 & 85.08 & 85.63 & 60.92 & 86.47 & 57.96 & 73.26 & 67.39 &    74.88 \\
    \hline
    
    PSPNet / ResNet-101 / Softmax & 94.01 & 68.08 & 88.80 & 64.87 & 75.87 & 95.60 & 89.59 & 93.15 & 37.96 & 88.20 & 72.58 & 89.96 & 93.30 & 87.52 & 86.65 & 61.90 & 87.05 & 60.81 & 87.13 & 74.65 &     80.63 \\
    
    PSPNet / ResNet-101 / SegSort & 96.00 & 67.17 & 93.37 & 74.52 & 77.77 & 95.07 & 89.39 & 93.91 & 41.31 & 87.85 & 73.66 & 90.15 & 91.06 & 85.63 & 87.86 & 71.81 & 90.28 & 65.99 & 86.75 & 75.53 &     82.41 \\
    
    \Xhline{1pt}
    \end{tabular}}
    \vspace{0.5pt}
    \caption{Per-class results on Pascal VOC 2012. The first 4 rows with gray colored background are on validation set while the last 4 rows are on testing set. Networks trained with SegSort consistently outperform their parametric counterpart (Softmax) by 1.63 to 2.43\%.}
    \label{tab:voc_per_class}
\end{table*}

\subsection{DeepLabv3+ / ResNet-101 Results on VOC}

We also train our supervised SegSort using DeepLabv3+ with ResNet-101 backbone with exactly same hyper-parameters as MNV2 backbone. We include the per-category segmentation performance in Table \ref{tab:voc_app}. Even though the hyper-parameters might not be optimal, we observe consistent performance improvements over the baseline Softmax method.

\begin{table*}
  \centering
  \resizebox{\textwidth}{!}{%
  \begin{tabular}{l|c c c c c c c c c c c c c c c c c c c c|c}
    \Xhline{1pt}
    Base / Backbone / Method & aero & bike & bird & boat & bottle & bus & car & cat & chair & cow & table & dog & horse & mbike & person & plant & sheep & sofa & train & tv & mIoU \\
    \hline \hline

    Deeplabv3$^+$ / ResNet-101 / Softmax & 90.93 & 56.70 & 89.46 & 73.35 & 82.13 & 95.03 & 87.30 & 91.88 & 37.79 & 83.56 & 56.32 & 88.31 & 83.32 & 86.11 & 86.61 & 58.17 & 87.65 & 52.87 & 88.43 & 74.19 &     78.24 \\
    
    Deeplabv3$^+$ / ResNet-101 / SegSort & 88.78 & 51.17 & 88.12 & 70.45 & 83.89 & 95.12 & 88.74 & 94.34 & 43.12 & 86.24 & 59.07 & 88.86 & 88.11 & 86.92 & 87.58 & 56.91 & 85.46 & 55.32 & 89.01 & 73.77 &     78.85 \\
    \hline

    

    \Xhline{1pt}
    \end{tabular}}
    \vspace{0.5pt}
    \caption{Per-class results on Pascal VOC 2012 validation set, using Deeplabv3$^+$ with Resnet-101 backbone.}
    \label{tab:voc_app}
\end{table*}

\end{document}